\documentclass[10pt,twocolumn,a4paper]{IEEEtran}

\usepackage[utf8]{inputenc} 
\usepackage[T1]{fontenc}    
\usepackage{hyperref}       
\usepackage{url}            
\usepackage{booktabs}       
\usepackage{amsfonts}       
\usepackage{nicefrac}       
\usepackage{microtype}      
\usepackage{amsmath}
\usepackage{bm}
\usepackage{bbm}
\usepackage{textcomp}
\usepackage{times}
\usepackage{xcolor}
\usepackage{graphicx}
\usepackage{psfrag}
\usepackage{ae,epsfig}
\usepackage{verbatim} 
\usepackage{enumerate} 
\usepackage{amsmath}
\usepackage{graphicx}
\usepackage{algorithm}
\usepackage{algpseudocode}

\usepackage{epstopdf}
\usepackage{tikz}
\usetikzlibrary{shapes,arrows}
\usetikzlibrary{positioning,fit}
\usepackage{color}
\usetikzlibrary{positioning}
\usepackage{caption}
\usepackage{subcaption}
\newcounter{teocount}
\newcounter{propcount}

\newcounter{remcount}
\newcounter{defcount}

\newcounter{excount}

\newcounter{lemcount}

\newtheorem{remm}[remcount]{Remark}

\newtheorem{definition}[defcount]{Definition}
\newtheorem{proposition}[propcount]{Proposition}
\newtheorem{theorem}[teocount]{Theorem}

\newtheorem{ex}[excount]{Example}

\newtheorem{lemma}[lemcount]{Lemma}

\newenvironment{proof}{{\em Proof. }}{\hfill \hspace*{1pt} \hfill $\blacksquare$}
\newenvironment{remark}{\begin{remm}\rm }{\hfill \hspace*{1pt} \hfill
$\star$\end{remm}}

\title{Reduced-Order Neural Network Synthesis with Robustness Guarantees}

\author{Ross Drummond, Matthew~C.~Turner and Stephen R. Duncan
\thanks{ Ross Drummond and Stephen R. Duncan are with the Department of Engineering Science, University of Oxford, OX1 3PJ, Oxford, United Kingdom. Email: \texttt{\{ross.drummond,stephen.duncan\}@ eng.ox.ac.uk.}

 Mathew~C.~Turner is with the Department of Electronics and Computer Science,
   University of Southampton, Southampton, UK, SO17 1BJ. Email:
   \texttt{m.c.turner@soton.ac.uk}
}
}

\begin{document}

\maketitle

\begin{abstract}

In the wake of the explosive growth in smartphones and cyberphysical systems,  there has been an accelerating shift in how data is generated away from centralised data towards on-device generated data. In response, machine learning algorithms are being adapted to run locally on board, potentially hardware limited, devices to improve user privacy, reduce latency and be more energy efficient. However, our understanding of how these device orientated algorithms behave and should be trained is still fairly limited. To address this issue, a method to automatically synthesize \emph{reduced-order} neural networks (having fewer neurons) approximating the input/output mapping of a larger one is introduced. The reduced-order neural network's weights and biases are  generated from a convex semi-definite programme that  minimises the worst-case approximation error with respect to the larger network. Worst-case bounds for this approximation error are obtained and the approach can be applied to a wide variety of neural networks architectures.  What differentiates the proposed approach to existing methods for generating small neural networks, e.g. pruning, is the inclusion of the worst-case approximation error directly within the training cost function, which should add robustness to out-of-sample data-points. Numerical examples highlight the potential of the proposed approach. The overriding goal of this paper is to generalise recent results in the robustness analysis of neural networks to a robust synthesis problem for their weights and biases. 



\end{abstract}

\section{Introduction}

As smartphones get increasingly integrated into our daily lives and the numbers of both cyberphysical systems and smart devices continues to grow, there has been a noticeable evolution in the way many large data sets are being generated.  In fact, Cisco \cite{cisco} predicted that in 2021, whilst 20.6 ZB of data (e.g. large ecommerce site records) will be handled by cloud-based approaches in large data-centres, this amount will be dwarfed by the 850 ZB  generated by local devices \cite{zhou}. In response to data sources becoming more device-centric, there has been a  shift in focus for many machine learning algorithms towards both implementing and training them locally on (potentially hardware limited) devices. Running the algorithms on the devices represents a radical shift away from traditional  \emph{centralised learning} where the data and algorithms are stored and processed in the cloud but, as described in \cite{zhou}, brings the benefits of  i) increased user privacy as the data is not transmitted to a centralised sever ii) reduced latency since the algorithms can react immediately to newly generated data from the device and iii) improved energy efficiency mostly because the data and algorithm outputs don't have to be constantly transferred to and from the cloud. However, running algorithms locally on devices brings its own issues, most notably in dealing with the devices' limited computational power, memory and energy storage. Overcoming these hardware constraints has motivated substantial efforts on improving algorithm design, particularly towards developing leaner, more efficient neural networks \cite{sze}.

Two popular approaches to make neural network algorithms leaner and more hardware-conscious are i) quantised neural networks \cite{quant1,quant2,quant3}, where fixed-point arithmetic is used to accelerate the computational speed and reduce memory footprint, and ii) pruned neural networks \cite{liu2018rethinking,prune_state,janowsky,karnin, mozer1,mozer2,lecun,hassibi,gale,lee,lin,han,molchanov,li2016}, where, typically, the weights contributing least to the function mapping are removed, promoting sparsity in the weights. Both of these approaches have achieved impressive results. For instance, by quantising, \cite{lin2016fixed}  was able to reduce model size by $>20\%$ without any noticeable loss in accuracy when evaluated on the CIFAR-10 benchmark and \cite{han2015learning} demonstrated that between 50-80$\%$ of its model weights could be pruned with little impact on performance \cite{sze}. However, our understanding of neural network reduction methods such as these remains lacking and reliably predicting their performance remains a challenge. Illustrating this point,  \cite{liu2018rethinking} stated that for pruned neural networks ``our results suggest the need for more careful baseline evaluations in future research on structured pruning methods'' with  a similar sentiment raised in \cite{prune_state}  ``our clearest finding is that the community suffers from a lack of standardized benchmarks and metrics''. These quotes indicate a need for robust evaluation methods for lean neural network designs, a perspective explored in this work. 

\subsubsection*{Contribution} 
\begin{figure}
\centering
\includegraphics[width=0.35\textwidth]{{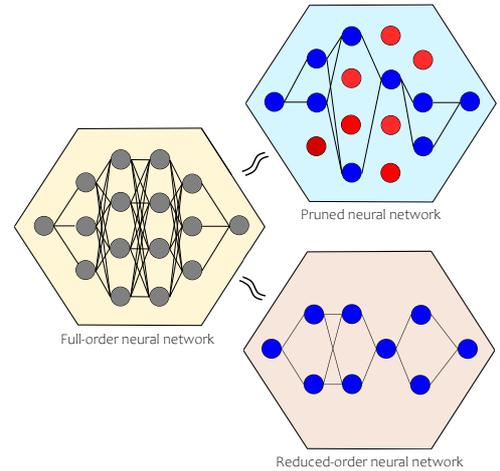}}
        \caption{Illustration of two different approximations of a neural network (termed the full-order network) to enable it to be run on limited hardware. One approach is to use network pruning to make the weights sparse while the second is to develop a reduced-order network with fewer neurons. This paper proposes a method to synthesize the weights and biases of the reduced order network such that they robustly minimise the approximation error with respect to the full order network. }
        \label{fig:comp}
\end{figure}

This paper introduces a method to automatically synthesize neural networks of reduced dimensions (meaning fewer neurons) from a trained larger one, as illustrated in Figure \ref{fig:comp}. These smaller networks are termed \emph{reduced-order neural networks} since the approach was inspired by reduced order modelling in control theory \cite{glover_modred}. The weights and biases of the reduced order network are generated from the solution of a semi-definite program (SDP)- a class of well-studied convex problems \cite{boyd} combining a linear cost function with a linear matrix inequality (LMI) constraint- which minimises the worst-case approximation error with respect to the larger network. Bounds are also obtained for this worst-case approximation error and so the performance of the network reduction is guaranteed. In this way, the method is said to be ``robust''  as it ensures the approximation error of the reduced-order neural network remains bounded for all input data in certain pre-defined sets, in a manner specified by the bound of Theorem \ref{thm:result}.

What separates the proposed synthesis approach to the existing methods for generating efficient neural networks, e.g. pruning, is the inclusion of the worst-case approximation error of the reduced-order neural network directly within the cost function for computing the weights and biases.  It is expected that this approach should offer two main advantages over classical pruning methods:
\begin{enumerate}
\item The method is robust in the sense that it provides guarantees of the approximation error with respect to the full-order network, unlike with pruning.
\item The method is one of the first to do automatic neural network \emph{synthesis} from the solution of a robust optimisation problem, with the weights and biases of the reduced-order neural networks generated in one-shot by solving a convex semi-definite program. Besides being of theoretical interest as an alternative to training via backpropogation, the main advantage of this approach is that it allows the worst-case error to be included directly within the training cost function which may result in out-of-sample generality in worst-case settings.
\end{enumerate}
Whilst the presented results are still preliminary, their focus on robust neural network synthesis introduces a new set of of tools to generate lean neural networks which should have more reliable out-of-sample performance, and which are equipped with approximation error bounds. The broader goals of this work are to translate recent results on the verification of NN robustness using SDPs \cite{pappas,raghunathan} into a synthesis problem, mimicking the progression from absolute stability theory \cite{zames} to robust control synthesis \cite{doyle} witnessed in control theory during the 1980s. In this way, this work carries on the tradition of control theorists exploring the connections between robust control theory and neural networks, as witnessed since the 1990s with Glover \cite{chu}, Barabanov \cite{barabanov}, Angeli \cite{angeli} and Narendra \cite{narendra}.

 \subsection{Notation}
 
  Non-negative real vectors of dimension $n$ are denoted $\mathbb{R}^n_+$. A positive (negative) definite matrix $\Omega$ is denoted $\Omega \succ (\prec )~ 0$. Non-negative diagonal matrices of dimension $n\times n$ are $\mathbb{D}^{n}_{+}$. The matrix of zeros of dimension $n \times m$ is $\bm{0}_{n \times m}$ and the vector of zeros of dimension $n$ is $\bm{0}_n$. The identity matrix of size $n$ is $I_n$. The vector of 1s of dimension $n$ is $\bm{1}_n$  and the $n \times m$ matrix of 1s is $\bm{1}_{n \times m}$. The $i^{\text{th}}$ element of a vector $x$ is denoted $x_i$ unless otherwise defined in the text. The $\star$ notation is adopted to represent symmetric matrices in a compact form, e.g.
 \begin{align}
 \begin{bmatrix}A & B \\ B^T & C \end{bmatrix} = \begin{bmatrix}A & B \\ \star & C \end{bmatrix}.
 \end{align}
 
\subsection{Neural networks}
 
 The neural networks considered will be treated as functions $f(x):\mathcal{X} \to \mathcal{F}$ mapping input vectors of size $ x \in \mathbb{R}^{n_x}$ to  output vectors of dimension $f(x) \in \mathbb{R}^{n_f}$. In a slight abuse of notation, $\phi(\cdot):\mathbb{R}^{n_k}\to \mathbb{R}^{n_k}$ will refer to mappings of both scalars and vectors, with the vector operation applied element-wise. The full-order neural network will be composed of $l$ hidden layers, with the $k^{\text{th}}$ layer being composed of $n_k$ neurons. The total number of neurons in the full-order neural network is $N = \sum_{k = 1}^{l} n_k$. 
 Similarly, the reduced-order neural network will be composed of $\lambda$ hidden layers with the $k^{\text{th}}$ layer being composed of $m_k$ neurons. The total number of neurons in the reduced-order network is $M = \sum_{k = 1}^{\lambda} m_k$.
 The dimension of the domain of the activation functions is defined as $\bar{N}=N-n_{l}+n_x$ (full-order network) and $\bar{M}=M-n_{\lambda}+n_x$ (reduced-order network).
 
 
 


 
 \section{Problem statement}
In this section, the general problem of synthesizing reduced-order NNs is posed. Consider a nonlinear function $f(x): \mathcal{X} \to \mathcal{F}$ mapping input data $x \in \mathcal{X}$ to an output set $\mathcal{F}$. The goal of this work is to generate a ``\textit{simpler}'' function $g(x): \mathcal{X} \to \mathcal{G}$ that is as ``\textit{close}'' as possible to $f(x)$ for all $x \in \mathcal{X}$. Here, ``\textit{simpler}'' will refer to the dimension of the reduced-order neural network's  weight matrix  being smaller than the full-order one and ``\textit{close}''-ness relates to the approximation error between the two functions $f(x)$ and $g(x)$ measured by the induced 2-norm $ \sup_{x \in \mathcal{X}} \|f(x)-g(x)\|_2$. 
The goal is to automatically synthesize the simpler functions $g(x)$ from the solution of a convex problem and obtain worst-case bounds for approximation error with respect to the larger neural network $f(x)$ for all  $x \in \mathcal{X}$.

To ensure that the function approximation problem remains feasible, structure is added to the set $\mathcal{F}$. It is assumed that the function being approximated $f(x)$ is generated by a feed-forward neural network 
\begin{subequations}\label{full}\begin{align}
x^0  & = x,
\\
x^{k+1}  & = \phi(W^kx^k + b^k), \quad k= 0, 1, \dots, l-1,
\\
f(x)  & = W^lx^l + b^l.
\end{align}\end{subequations}
Here, the input data $x^0 = x \in \mathcal{X} \subseteq \mathbb{R}^{n_x}$ is mapped through the nonlinear activation functions $\phi(\cdot)$ (which could be the standard choices of ReLU, sigmoid, tanh or any function that satisfies a quadratic constraint as given in Section \ref{sec:QI_act}) element-wise with the weight matrices $W^0 \in \mathbb{R}^{n_1 \times n_x}$, $W^k \in \mathbb{R}^{n_{k+1} \times n_k}$ and biases $b^k \in \mathbb{R}^{n_{k+1}}$, $ k = 0,\, \dots, l-1$.
Whilst the results are described for feed-forward neural networks, the method can be generalised to other network architectures, such as recurrent and even implicit neural networks \cite{implicit}. As an aside, verifying the well-posedness of implicit neural networks has a strong connection to that of Lurie systems with feed-through terms \cite{valmorbida}.  

{
The network (\ref{full}) can be equivalently written in the implicit form
\begin{subequations}\label{full2}\begin{align}
\check{x} & = \phi(W \check{x} + W_0 x + b), \quad \quad \phi(.):~ \mathbb{R}^N \mapsto \mathbb{R}^N,
\\
f(x) & = W^f \check{x} + b^l ,
\end{align}\end{subequations}
where
\begin{subequations} \begin{align}
  \check{x} & =
  \begin{bmatrix}
   x^1 \\ x^2 \\ \vdots \\ x^l 
  \end{bmatrix},\, 
  W = 
  \begin{bmatrix}
    0   & 0 & \hdots & 0  \\
    W^1 & \ddots & \ddots &  \vdots \\
    \vdots & \ddots &0  & 0  \\
    0  & \dots &  W^{l-1} & 0 
   \end{bmatrix},\,
   W_0 = \begin{bmatrix}
   W^0 \\ 0 \\ \vdots \\ 0 
  \end{bmatrix},\, \\
  b & = \begin{bmatrix}
   b^0 \\ b^1 \\ \vdots \\ b^{l-1} 
  \end{bmatrix}, \quad
  W^f = \begin{bmatrix} 
       0, &  \hdots\,,& 0, & W^{l} 
      \end{bmatrix}.
 \end{align}
 \end{subequations}
 }
The neural network $f(x)$ (which will be referred to as the \textit{full-order neural network}) is to be approximated by another neural network $g(x)\in \mathcal{G}$ (referred to as the \textit{reduced-order neural network}) of a smaller dimension
\begin{subequations}\label{reduced}\begin{align}
z^0  & = x,
\\
z^{k+1}  & = \phi\left(\sum_{i =0}^{\lambda-1}\Psi^{k+1,i}z^i + \beta^k\right), \quad k= 0, 1, \dots, \lambda-1, \label{z_update}
\\
g(x)  & = \Psi^{\lambda}z^{\lambda} + \beta^{\lambda}.
\end{align}\end{subequations}
The weights and biases in this neural network are $\Psi^{k,i} \in \mathbb{R}^{m_{k+1} \times m_i}$,  $\beta^k \in \mathbb{R}^{m_{k+1}}$, $k = 0, \, \dots, \, \lambda-1$. The network structure in \eqref{z_update} is general, and even allows for implicitly defined networks \cite{implicit}. This generality follows from the lack of structure imposed on the matrices used in the synthesis procedure. However, by adding structure, the search can be limited to, for example, feed-forward networks, which are simpler to implement.  

{
Similar to the full-order case, the network (\ref{reduced}) can be written as
\begin{subequations}\label{reduced2}
\begin{align}
  \check{z} & =   \phi ({\Psi} \check{z} + \Psi_0 x + \beta ), \quad \phi(\cdot): \mathbb{R}^M \mapsto \mathbb{R}^M, \\
  g         & =   \Psi^f \check{z} + \beta^{\lambda} ,
 \end{align}
\end{subequations}
where
\begin{subequations}
\begin{align}
\check{z} & = \begin{bmatrix}
   z^1 \\ z^2 \\ \vdots \\ z^\lambda 
  \end{bmatrix}, \,
 \Psi =  \begin{bmatrix}
    \Psi^{1,1}   & \Psi^{1,2} & \hdots & \Psi^{1,\lambda}  \\
     \Psi^{2,1} & \Psi^{2,2} & \ddots & 
     \vdots\\
     \vdots & \ddots & \ddots  & \Psi^{\lambda-1,\lambda} \\
     \Psi^{\lambda,1}  & \dots & \Psi^{\lambda,\lambda-1} & \Psi^{\lambda,\lambda} 
    \end{bmatrix} ,\,
    \\
    \Psi_0  & = \begin{bmatrix}
              \Psi^{1,0} \\
              \Psi^{2,0} \\
              \vdots 
              \\ \Psi^{\lambda,0}
             \end{bmatrix} ,\,
 \beta  = \begin{bmatrix}
   \beta^0 \\ \beta^1 \\ \vdots \\ \beta^{\lambda-1}     \end{bmatrix},\,
   \Psi^f = \begin{bmatrix} 
       0, &  \hdots\,, &0, & \Psi^{\lambda} 
       \end{bmatrix}.
\end{align}
\end{subequations}
}

In this work, the dimension of the reduced-order network is fixed and the problem is to find the reduced-order NN's parameters, being the weights $\Psi^{k,i}$ and biases  $\beta^k$, that minimise  the worst-case approximation error between the full and reduced order neural networks for all $x \in \mathcal{X}$. In practice, the dimension of the reduced-order network should be fixed to the minimum value for which Proposition \ref{prop:result} can be solved to give a sufficient level of performance, as typically increasing the dimension of this neural network should lead to improved approximations to the full-order one, as larger networks will be more expressive allowing them to more accurately approximate highly nonlinear functions.  The main tool used for this reduced-order NN synthesis problem is the outer approximation of the NN's input set $\mathcal{X}$, nonlinear activation function's gains $\phi(\cdot)$ and the output error $f(x)-g(x)$ by quadratic constraints.  These outer approximations enable the robust weight synthesis problem to be stated as a convex SDP, albeit at the expense of introducing conservatism into the analysis. 

\section{Quadratic Constraints} 
In this section, the quadratic constraints for the convex outer approximations of the various sets of interest of the reduced NN synthesis problem are defined. These characterisations are posed in the framework of \cite{pappas}, which in turn was inspired by the integral quadratic constraint framework of \cite{iqc} and the classical absolute stability results for Lurie systems \cite{khalil}.

\subsection{Quadratic constraint: Input set}

The input data $x \in \mathcal{X}$ is restricted to the hyper-rectangle $\mathcal{X}_{\infty}$. 
\begin{definition}\label{def:x_gen}
Define the hyper-rectangle  $\mathcal{X}_{\infty}= \{x: \underline{x}_i \leq x_i \leq \overline{x}_i,\, i = 1, \, \dots, \, n_x\}$. 
If $x \in \mathcal{X}_{\infty}$ then 
$
[x^T ~ 1 ]P_{\mathcal{X}_{\infty}}[x^T ~ 1 ]^T \geq  0 
$
where 
\begin{align}
P_{\mathcal{X}_{\infty}} = \begin{bmatrix}  -\tau_{x_{\infty}}& \frac{\tau_{x_{\infty}}}{2}\left(\underline{x} + \overline{x}\right)  
\\ 
\star &  -\underline{x}^T\tau_{x_\infty}\overline{x} \end{bmatrix}, \quad \tau_{x_{\infty}}\in \mathbb{D}^{n_x}_{+}.
\end{align}

{
Note that the input set constraint characterised by Definition \ref{def:x_gen} can be equivalently written
as
\begin{align} \label{eq:ip}
  \omega(x)^T \Pi_{\infty} \omega(x) \geq 0
\end{align}
where 
\[
 \Pi_{\infty} =
 \begin{bmatrix}  -\tau_{x_{\infty}}& 0_{n_x \times N} & 0_{n_x \times M} & \frac{\tau_{x_{\infty}}}{2}\left(\underline{x} + \overline{x}\right)  
\\ 
\star & 0_{N \times N} & 0_{N \times M} & 0_{N \times 1} \\
\star & \star & 0_{M \times M} & 0_{M \times 1} \\
\star & \star& \star &  -\underline{x}^T\tau_{x_\infty}\overline{x} \end{bmatrix}
\]
and
\begin{align}
\mu(x) = \begin{bmatrix} x^T & \check{x}^T & \check{z}^T & 1\end{bmatrix}^T.
\end{align}
}
\end{definition}

\subsection{Quadratic constraint: Activation functions}\label{sec:QI_act}

The main obstacle to any robustness-type result for neural networks is accounting for the nonlinear activation functions $\phi(\cdot)$. To address this issue, the following function properties are introduced.

\begin{definition}\label{def:prop}
The activation function $\phi(s): \mathcal{S} \subset \mathbb{R} \to \mathbb{R}$ satisfying $\phi(0) = 0$ is said to be \emph{sector bounded} if 
\begin{subequations}\begin{align}
\frac{\phi(s)}{s} \in [0, \delta] \quad \forall s \in \mathcal{S}, \quad \delta > 0,
\end{align}
and \emph{slope restricted} if 
\begin{align}
\frac{d\phi(s)}{ds} \in [\underline{\beta}, \beta],  \quad \forall s \in \mathcal{S}, \quad \beta >0.
\end{align}
If $\underline{\beta} = 0$ then the nonlinearity is \emph{monotonic} and if $\phi(s)$ is slope restricted then it is also sector bounded. The activation function $\phi(s)$ is \emph{bounded} if 
\begin{align}
\phi(s) \in [\underline{c}, \overline{c}],\quad \forall s \in \mathcal{S},
\end{align}
it is \emph{positive} if
\begin{align}
\phi(s) \geq 0 , \quad \forall s \in \mathcal{S},
\end{align}
its \emph{complement is positive} if
\begin{align}
\phi(s)-s \geq 0 , \quad \forall s \in \mathcal{S},
\end{align}
and it satisfies the \emph{complementarity condition} if 
\begin{align}
(\phi(s)-s)\phi(s) = 0, \quad \forall s \in \mathcal{S}.
\end{align}\end{subequations}
\end{definition}
Most popular activation functions, including the ReLU, (shifted-)sigmoid and tanh satisfy one or more of these conditions, as illustrated in Table \ref{tab:comparison}.  As the number of properties satisfied by $\phi(\cdot)$ increases, the characterisation of this function  within the robustness analysis improves, often resulting in less conservative results. It is also noted that to satisfy $\phi(0) = 0$ some activation functions may require a shift, e.g. the sigmoid, or they may require transformations to satisfy additional function properties, as demonstrated in the representation of the LeakyReLU as a ReLU + linear term function.

\begin{table*}[!htb]
 \centering
 \begin{tabular}{|l|l|l|l|l|}
 \hline
  $\phi(\cdot)$ property & Shifted sigmoid & tanh & ReLU  & ELU \\ \hline
  Sector bounded      &  \checkmark &  \checkmark &  \checkmark  & \checkmark \\
  Slope restricted      &  \checkmark &  \checkmark & \ \checkmark & \checkmark \\
  Bounded & \checkmark&  \checkmark & \texttimes  & \texttimes  \\
  Positive & \texttimes & \texttimes & \checkmark  & \texttimes  \\
  Positive complement &\texttimes  &\texttimes & \checkmark  & \texttimes  \\
  Complementarity condition         & \texttimes   & \texttimes  &   \checkmark & \texttimes  \\
   \hline
 \end{tabular}
 \caption{\label{tab:comparison} Properties of commonly used activation functions, including the sigmoid, tanh, rectified linear unit ReLU and exponential linear unit (ELU). The properties of other functions, such as the LeakyReLU, can also be inferred. }
\end{table*}

As is well-known from control theory \cite{khalil}, functions with these specific properties are important for robustness analysis problems because they can be characterised by quadratic constraints. 
\begin{lemma}\label{lem:prop}
Consider the vectors $y,y_1 \in \mathbb{R}^{n_y}$,  and $\upsilon \in \mathbb{R}^{n_{\upsilon}}$ that are mapped component-wise through the activation functions $\phi(\cdot):\mathbb{R}^{n_y} \to \mathbb{R}^{n_y}$ and $\tilde{\phi}(\cdot):\mathbb{R}^{n_v} \to \mathbb{R}^{n_v}$. If $\phi(y)$ is
\emph{sector-bounded}, then
\begin{subequations}\label{qi_gen}\begin{align}
 (\delta y-\phi(y))^T{\bf T}^{\text{s}}\phi(y)  & \geq 0,\quad \forall y \in \mathbb{R}^{n_y}, ~{\bf T}^{\text{s}}  \in \mathbb{D}^{n_y}_+ ; \label{sec_quad}
\end{align}
\emph{slope-restricted} then
 \begin{equation*}
\resizebox{0.48\textwidth}{!}{$(\beta (y-y_1)-(\phi(y)-\phi(y_1))^T{\bf T}^{\text{sl}} (\phi(y)-\phi(y_1)-\underline{\beta}(y-y_1))  \geq 0;
$}
 \end{equation*} 
 \vspace{-0.5cm}
 \begin{equation}
     \quad \quad \quad  \forall \{y, \, y_1 \} \in \mathbb{R}^{n_y},\,  {\bf T}^{\text{sl}}  \in \mathbb{D}^{n_y}_+;
 \end{equation}
\emph{bounded} then 
\begin{align}
 (\overline{c}-\phi(y))^T{\bf T}^{B}(\phi(y)-\underline{c})  & \geq 0,\quad \forall y \in \mathbb{R}^{n_y}, \, {\bf T}^{B}\in \mathbb{D}^{n_y}_+;
\end{align}
\emph{positive} then
\begin{align}
({\bf T}^{+})^T  \phi(y)\geq 0,  \quad \forall y \in \mathbb{R}^{n_y},\, {\bf T}^{+} \in \mathbb{R}_+^{n_y};
\end{align}
such that is \emph{complement is positive} then 
\begin{align}
({\bf T}^{c+})^T (\phi(y)-y) \geq 0,  \quad \forall y \in \mathbb{R}^{n_y}, \, {\bf T}^{c+}  \in \mathbb{R}^{n_y}_+.
\end{align}
If $\phi(y)$ satisfies the \emph{complementary} condition then 
\begin{align}
 (\phi(y)-y)^T{\bf T}^{0}\phi(y)  = 0,\quad \forall y \in \mathbb{R}^{n_y},\, {\bf T}^{0} \in \mathbb{D}^{n_y}.\label{comp_quad}
\end{align}
Additionally, if both $\phi(y)$ and $\tilde{\phi}(\upsilon)$ and their complements are positive then so are the \emph{cross terms} 
\begin{align}\label{cross_qcs_1}
\small
 \tilde{\phi}(\upsilon)^T{\bf T}^{\times}(\phi(y)-y) \geq 0,  \; \forall \upsilon \in \mathbb{R}^{n_v}, y \in \mathbb{R}^{n_y}, \, {\bf T}^{\times} \in \mathbb{R}_+^{n_{\upsilon} \times n_y}, \\
 \tilde{\phi}(\upsilon)^T{\bf T}^{\otimes}\phi(y) \geq 0,  \; \forall \upsilon \in \mathbb{R}^{n_v}, y \in \mathbb{R}^{n_y}, \, {\bf T}^{\otimes} \in \mathbb{R}_+^{n_{\upsilon} \times n_y}  .\label{cross_qcs_2}
\end{align}\end{subequations}
\end{lemma}

Inequalities \eqref{sec_quad}-\eqref{comp_quad} are well-known however the cross terms \eqref{cross_qcs_1}-\eqref{cross_qcs_2} acting jointly on activation function pairs are less so. 

{
\begin{remark}
Lemma \ref{lem:prop} is established globally, that is for all $y \in \mathbb{R}^{n_y}$. Some activation functions $\phi(\cdot)$ may be defined locally, or the sector, slope bounds may be tighter for restricted values of their arguments. In such cases, local versions of Lemma \ref{lem:prop} may give less conservative results.   
\end{remark}
}

The characterisation of the nonlinear activation functions via quadratic constraints allows the neural network robustness analysis to be posed as a SDP- with the various $\lambda$'s in Lemma \ref{lem:prop} being decision variables.  Such an approach has been used in \cite{pappas, glover_modred, barabanov}, and elsewhere, for neural networks robustness problems, with the conservatism of this approach coming from the obtained worst-case bounds holding for all nonlinearities satisfying the quadratic constraints. In this work, the aim is to extend this quadratic constraint framework for neural network robustness analysis problems to a synthesis problem.


A quadratic constraint characterisation of both the reduced and full-order neural networks can then be written, with the following lemma being the application of Lemma \ref{lem:prop} for both the reduced and full-order neural networks. 
 
 {
\begin{lemma}\label{ass_phi}
If the activation function $\phi(\cdot)$ satisfies one or more of the quadratic constraints of Lemma \ref{lem:prop}, then there exists a matrix 
\begin{align}
\Lambda = \begin{bmatrix} \bm{0}_{n_x \times n_x}&    \Lambda_{12} &  \Lambda_{13} &  \Lambda_{14} 
\\
\star &   \Lambda_{22} &  \Lambda_{23} &  \Lambda_{24} 
\\
\star &  \star &  \Lambda_{33} &  \Lambda_{34} 
\\
\star &  \star &  \star &  \Lambda_{44}  
\end{bmatrix}    ,
\end{align}
defined by the ${\bf T}^i$'s ($i \in \left\{\text{s},\text{sl},+,c+,B,0,\times,\otimes \right\}$) of Lemma \ref{lem:prop} such that 
\begin{align}
\mu(x)^T \Lambda \mu(x)
 \geq 0, \quad \forall x \in \mathcal{X}.\label{eq_ass_phi}
\end{align}
\end{lemma}
}

{
\proof{The construction of $\Lambda$ for the sector nonlinearity associated with the full-order network is shown. Lemma \ref{lem:prop} implies that for a matrix ${\bf T}^{\text{s}} \in \mathbb{R}^{N \times N}$
\[
2 \phi(\xi)^T {\bf T}^{\text{s}} (\xi - \phi(\xi)) \geq 0
\]
where, from equation (\ref{full2}), $\xi = W \check{x} + W_0 x + b$. Noting that $\phi(\xi) = \check{x}$, expanding the above becomes
\[
 2 \check{x}^T {\bf T}^{\text{s}} (W \check{x} + W_0 x + b - \check{x}) \geq 0
\]
and majorising it gives
\begin{align}
 \begin{bmatrix}
  x \\ \check{x} \\ \check{z} \\ 1 
 \end{bmatrix}^T
  \begin{bmatrix}
   0 & W_0^T {\bf T}^{\text{s}} & 0 & 0 
   \\
   \star & -2 {\bf T}^{\text{s}} + {\bf T}^{\text{s}} W + W^T {\bf T}^{\text{s}} & 0 & {\bf T}^{\text{s}} b \\
   \star & \star & 0 & 0 \\
   \star & \star & \star & 0 
  \end{bmatrix}
\begin{bmatrix}
  x \\ \check{x} \\ \check{z} \\ 1 
 \end{bmatrix} \geq 0.
\end{align}
This clearly takes the form of the inequality in \eqref{eq_ass_phi}. All other cases are derived similarly. 
 }
 }
 Appendix 1 details the characterisation of $\Lambda$ for the specific case of the ReLU activation functions.


\subsection{Quadratic constraint: Approximation error of the reduced-order neural network}
An upper bound for the approximation error between the full and reduced-order networks can also be expressed as a quadratic constraint. This error bound will be used as a performance metric to gauge how well the reduced-order neural network approximates the full-order one, as in how well $g(x) \approx f(x)~ \forall x \in \mathcal{X}_{\infty}$. 

\begin{definition}[Approximation error]\label{def:error}
For some $\gamma_x \geq 0$, $\gamma \geq 0$, the reduced-order NN's approximation error  is defined as the quadratic bound
\begin{align}\label{approx_bound}
\|f(x)-g(x)\|_2^2 \leq \gamma_x \|x\|_2^2 + \gamma, \quad \forall x \in \mathcal{X}_{\infty}.
\end{align}
\end{definition}
In practice, this bound is computed by minimising over some weighted sum of $\gamma_x$ and $\gamma.$

Note that by using equations (\ref{full2}) and (\ref{reduced2})  the approximation error $f(x)-g(x)$ can be written as 
\begin{align}
f(x)-g(x) = L\mu(x)
\end{align}
where
\begin{align}\label{L} \small
L = 
\begin{bmatrix} 
0_{n_f \times n_x}, & W^f, & -\Psi^f, & b^l-\beta^{\lambda} 
\end{bmatrix}
\end{align}
{Similarly, 
\begin{align}
 \gamma_x \| x \|_2^2 + \gamma = \mu(x)^T \Gamma \mu (x)
\end{align}
where $\Gamma = {\rm blockdiag}(\gamma_x I_{n_x}, 0_{N \times N}, 0_{M \times M}, \gamma)$, so inequality (\ref{approx_bound}) is equivalent to 
\[
 \mu(x)^T (L^TL-\Gamma) \mu(x) \leq 0.
\]
}


\section{Reduced-order neural network synthesis problem}\label{sec:synth}
 
This section contains the main result of the paper; an SDP formulation of the reduced-order NN synthesis problem (Proposition \ref{prop:result}). To arrive at this formulation, a general statement of the synthesis problem is first defined in Theorem \ref{thm:result}. This theorem characterises the search for the reduced-order neural network's parameters as minimising the worst-case approximation error for all inputs $x \in \mathcal{X}$.

 \begin{theorem}\label{thm:result}
Assume the activation functions $\phi$ satisfy one or more of the properties from Definition \ref{def:prop}. With fixed weights $\{w_1, \, w_2\} \geq 0$, if there exists a solution to 
\begin{subequations} \begin{align}
   \min_{\Psi,\, \Psi_0, \, \Psi^{l},\,\beta,\, \beta^{\lambda},\,{\bf T}^i,\, {\bf T}_r^i, \tau_{x_\infty},\, \gamma_x, \, \gamma}  & ~  w_1\gamma_x + w_2\gamma , 
 \\
& \hspace{-3.8cm}\text{s.t. } \nonumber \\
& \hspace{-4.2cm} \mu(x)^T \label{qi_thm} \Pi_{\infty}\mu(x)  + \mu(x)^T \Lambda \mu(x)
 + \mu(x)^T(L^T L - \Gamma) \mu(x) \leq 0,
 \\ &  \hspace{-0.8cm} \gamma_x \geq 0, \, \gamma \geq 0, \nonumber
 \end{align}\end{subequations}
 then the worst-case approximation error is bounded by $\|f(x)-g(x)\|_2^2~\leq~\gamma_x~\|x\|_2^2~+~\gamma$ for all $x \in \mathcal{X}_{\infty}$.
 %
 %
 \end{theorem}
\proof{
See Appendix 3.}

The main issue with Theorem \ref{thm:result} is verifying inequality \eqref{qi_thm} since it includes a non-convex bilinear matrix inequality (BMI) between the matrix variables of the reduced-order network's weights,  its biases and the scaling variables in $\Lambda$. The following proposition details how this constraint can be written (after the application of a convex relaxation of the underlying BMI) as an LMI. The search over the reduced NN variables can then be translated into a SDP, a class of well understood convex optimisation problems with many standard solvers such as MOSEK \cite{mosek} implemented through the YALMIP \cite{yalmip} interface in MATLAB or even the Robust Control Toolbox. 

  \begin{proposition}\label{prop:result}
  Consider the full-order neural network of \eqref{full} mapping $x \to f(x)$ and the reduced-order neural network of \eqref{reduced} mapping $x\to g(x)$. For fixed weights $\{w_1,\, w_2\} \geq 0$, if there exists matrix variables ${\bf T}^i$ (of appropriate dimension and property\footnote{From Lemma \ref{lem:prop} the matrices $\mathbf{T}^i$ and $\mathbf{T}_r^i$ may have special properties such that they must have positive elements or be diagonal.}), ${\bf F}_\Psi\in \mathbb{R}^{M \times {M}},\,{{\bf F}_0}\in \mathbb{R}^{M \times n_x},\, {\bf F}_\beta\in \mathbb{R}^{M}$, $\Psi^f \in \mathbb{R}^{n_f \times M}$ and $\beta^{\lambda}\in \mathbb{R}^{n_f}$ that solve
  \begin{subequations}\label{omega}\begin{align} 
   & \min _{{\bf F}_\Psi,\,{\bf F}_0,\, {\bf F}_\beta, \, \Psi^f, \, \beta^{\lambda},\,{\bf T}^i,\,{\bf T}_r^i, \tau_{x_\infty},\, \gamma_x, \, \gamma} ~ w_1\gamma_x + w_2\gamma, 
  \\
  & \hspace{1.6cm}\text{s.t. }  \quad \Omega_{\text{Schur}} \prec 0, \, \gamma_x \geq 0, \, \gamma \geq 0, \label{LMI_prop}
  \end{align}\end{subequations}
  with $\Omega_{\text{Schur}}$ defined in \eqref{eq:schur} of Appendix 3, then the reduced-order network with weights and affine terms defined by
  \begin{align}
  \Psi_0 = U_1^{-1} {\bf F}_0 \quad \Psi = U_2^{-1} {\bf F}_\Psi  \quad \beta = U_3^{-1}{\bf F}_\beta, 
  \end{align}
  ensures that the worst-case approximation error bound of the reduced-order neural network satisfies $\|f(x)-g(x)\|_2^2~\leq~\gamma_x~\|x\|_2^2~+~\gamma$  for all $x \in \mathcal{X}_{\infty}$. 
  %
  
  \end{proposition}
  \proof{ See Appendix 3.}

Appendix 4 details how the matrix $\Lambda$, which characterises how the activation functions are included within the robustness condition $\Omega_{\text{Schur}} \succ 0$, can be written as a linear function of the matrix variables as required by Proposition \ref{prop:result} for the special case where $\phi(y) =\text{ReLU}(y)$.
Some remarks about the proposition are given in Appendix 5. 

\begin{remark}
There is some degree of flexibility in choosing the architecture of the reduced-order neural network in Proposition \ref{prop:result}. This flexibility is viewed as an advantage of the method, as it increases its applicability, but it is also acknowledged that it could make finding the optimal architecture more challenging. In practice, it has been observed that a suitable way to fix the architecture is to set the activation function to be the same as that of the full-order network and also to set the layer dimensions of the reduced-order network to be quite low. Upon solving Proposition 1, the user should then inspect the performance of the reduced-order network generated by Proposition 1. If a satisfactory level of performance has been achieved (measured either through the bounds or from inspecting the error to the full-order network directly), then this architecture should be retained. If not, then the dimension of the reduced-order network should be increased, and Proposition 1 ran again. This process should be repeated until the performance standards have been met. Algorithm \ref{alg} shows pseudo-code for this design process. 
\end{remark}

\begin{remark}
The reduced-order neural network generated by Proposition \ref{prop:result} could also be fine-tuned, as often applied to  pruned neural networks. 
\end{remark}

\begin{algorithm}[t]
\caption{Update the reduced-order neural network architecture}\label{alg}
\begin{algorithmic}
\Require Full-order neural network parameters ($W, b$), reduced-order neural network dimensions ($m_k$, $\lambda$) and tolerances  $\varepsilon_1$, $\varepsilon_2$.
\For{$j = 1, \,2,\, \dots, \, \mathcal{J}$}
\State $[\Psi,\beta] \gets$ Proposition \ref{prop:result}($m_k,\,\lambda$)
\State $p_j = \gamma_x~\|x\|_2^2+ \gamma $
\State $q_j = \max~\|f(x)-g(x)\|_2^2~\forall x \in \mathcal{X}_{\infty}' \subseteq \mathcal{X}_{\infty} $
\If {$p_j \leq \varepsilon_1$ and/or $q_j \leq \varepsilon_2$  } 
 \State Break
 \Else
\State Increase $m_k$ {and/or} $\lambda$.
\EndIf
\EndFor
\end{algorithmic}
\end{algorithm}


\section{Numerical example}\label{sec:examples} 
 
\begin{figure}[h!]
   \centering
\begin{subfigure}{.4\textwidth}
  \centering
  \includegraphics[width=.9\linewidth]{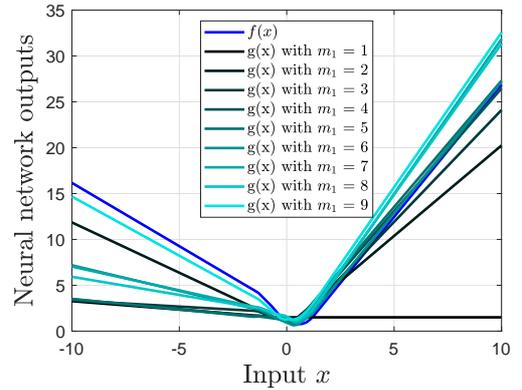}  
  \caption{Outputs of the full-order $f(x)$ (blue) and reduced-order $g(x)$ (grey) neural networks of various dimensions $m_1$. }
  \label{fig:functions}
\end{subfigure}
\quad
\begin{subfigure}{.4\textwidth}
  \centering
  \includegraphics[width=.9\linewidth]{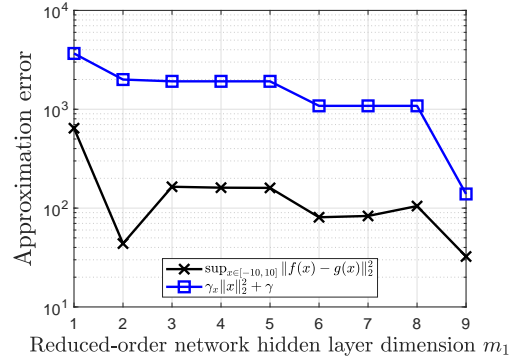}  
  \caption{Error bounds as a function of $m_1$.  The worst-case error between the full and reduced order neural networks is shown in black whilst the bounds obtained from Proposition \ref{prop:result} is shown in blue.}
  \label{fig:error}
\end{subfigure}
\caption{Evaluation of the reduced-order neural networks synthesized from Proposition \ref{prop:result} in the first numerical example.}
\label{fig:ex1}
\end{figure}

\begin{figure}
\centering
\includegraphics[width=0.4\textwidth]{{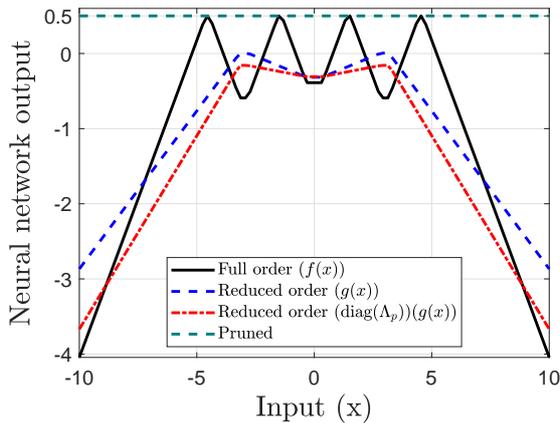}}
        \caption{Outputs of the second numerical example comparing the reduced order neural networks (with both full and diagonal $\Lambda_p$ matrices in Lemma \ref{ass_phi}) and a non-fine tuned pruned neural network of an equivalent size (with the 32 smallest weights being set to zero). With this level of reduction, the pruned network gave a constant output of $1/2$ whereas the reduced-order network could capture some of the variation of the function.  }
        \label{fig:ex2}
\end{figure}

 The proposed reduced-order neural network synthesis method was then evaluated in two numerical examples. In both cases, the performance of the synthesized neural networks were evaluated graphically (see Figures \ref{fig:ex1}-\ref{fig:ex2}) to give a better representation of the robustness of the approximations (the focus of this work). Only academic examples were considered due to the well-known scalability issues of SDP solvers (but which are becoming less of an issue \cite{dathathri2020enabling}) and because performance was measured graphically. The code for the numerical examples can be obtained on request from the authors.  
 
 The first example explores the impact of reducing the dimension of the reduced-order neural network on its accuracy. In this case, the full-order neural network considered was a single hidden layer network of dimension 10 with the weights $W^0$, $W^1$, $b^0$ and $b^l$ all obtained from sampling a zero mean normal distribution with variance 1 and which mapped a single input to a single output, $n_x = n_f = 1$, with the input constrained to $x \in [-10,10]$. The ReLU was taken as the activation function of both the full and reduced-order neural networks. Reduced-order neural networks with single hidden layers of various dimensions $m_1$ were then synthesized using Proposition \ref{prop:result}. Figure \ref{fig:functions} shows the various approximations obtained and Figure \ref{fig:error} shows how the error bounds and approximation errors changed as the dimension of the reduced-order network $m_1$ increased. The error bound was satisfied in all cases (albeit conservatively) and dropped as the degree of the reduced-order network increased. Even though the error bound (the blue line in the figure) from the solution of Proposition \ref{prop:result} monotonically decreased as the dimension of the reduced-order neural network increased, this did not imply that the actual observed worst-case error (black line in the figure)  would also monotonically decrease, as observed at $m_1 =2$. The non-monotonicity of the error highlights how the performance of several candidates reduced-order neural network architectures should be evaluated prior to deployment before an ``optimal’’ architecture is implemented, with Algorithm \ref{alg} illustrating one approach to conduct this search

The second example considers a more complex function to approximate and illustrates some potential pitfalls of pruning too hard. In this case, the full-order network's weights were defined by $\ell = 4$, $n_k = 4$, $n_x = 1$ and with $v~=~(1,\,\dots,\, n_k+1)/n_k$ then the weights and biases were
$    W^0~=~\cos\left(2\pi v\right)$,
$b_0~= 0$, 
    $W^k~=~\frac{1}{k+1}\left( \cos\left(2\pi v\right)\times  \sin\left(2\pi v\right)^T\right)$, $b^k = \frac{1}{k+1}\left(  \sin\left(2\pi v\right)\right)$, $W^\ell = \sin\left(2\pi v\right)$ and $b^\ell = 0.$ Figure \ref{fig:ex2} shows the output generated from a $\lambda = m_k = 3$ reduced-order neural network as well as the network generated by setting the $\Lambda_p$ matrices in Lemma \ref{ass_phi} to be diagonal (this reduced the compute time but, as shown, can alter the obtained function). Also shown is the case when the full-order neural network has been pruned to have a similar number of connections as the reduced-order one by removing the 32 smallest (out of a total of 56) weights. In this case, the pruned network was cut so far that it simply generated a constant function, but further fine-tuning of the pruned network may recover performance. Likewise, fine-tuning of the reduced-order neural network (through different substitutions of $J_1$ and $J_2$ or from simply applying the standard fine-tuning update of pruning) may improve the approximation of the synthesized reduced-order neural networks.

 \section*{Conclusions}
A method to synthesize the weights and biases of reduced-order neural networks (having few neurons) approximating the input/output mapping of a larger was introduced. A semi-definite program was defined for this synthesis problem that directly minimised the worst-case approximation error of the reduced-order network with respect to the larger one, with this error being bounded. By including the worst-case approximation error directly within the training cost function, it is hoped that the ideas explored in this paper will lead to more robust and reliable reduced-order neural network approximations. Several open problems still remain to be explored, most notably in reducing the conservatism of the bounds, scaling up the method to large neural networks and exploring the convexification of the bi-linear matrix inequality of the synthesis problem.

 \section*{Acknowledgements}
The authors were funded for this work through the Nextrode Project of the Faraday Institution (EPSRC Grant 	EP/M009521/1) and a UK Intelligence community fellowship from the Royal Academy of Engineering. 
 

\bibliographystyle{IEEEtranS}
\bibliography{bibliog_rnn}


 \section*{Appendices}
 
 \subsection*{Appendix 1: Quadratic Constraint for the ReLU}
 
 Consider the generalised quadratic constraint of Lemma \ref{ass_phi} with the $\phi(y) = $ReLU$(y)$ activation function and define both $\xi = W \check{x} + W_0 x + b$ and $\zeta = {\Psi} \check{z} + \Psi_0 x + \beta$.
{From Table \ref{tab:comparison}, it is clear that the \emph{full order} network activation function satisfies the following quadratic constraints:
 \begin{align}
  & 2 \phi(\xi)^T {\bf T}^{0} (\xi -\phi(\xi)) \geq 0, && {\bf T}^{0} \in \mathbb{R}^{N \times N}, \\
  & 2 ({\bf T}^{+})^T \phi (\xi) \geq 0, && {\bf T}^{+} \in \mathbb{R}_+^N, \\
  & 2 ({\bf T}^{c+})^T (\phi (\xi) -\xi) \geq 0, && {\bf T}^{c+} \in \mathbb{R}_+^N .
 \end{align}
 Similarly, the reduced order activation functions satisfy the following quadratic constraints
 \begin{align}
  & 2 \phi(\zeta)^T {\bf T}_r^{0} (\zeta -\phi(\zeta)) \geq 0, && {\bf T}_r^{0} \in \mathbb{R}^{M \times M}, \\
  & 2 ({\bf T}_r^{+})^T \phi (\zeta) \geq 0, && {\bf T}_r^{+} \in \mathbb{R}_+^M, \\
  & 2 ({\bf T}_r^{c+})^T (\phi (\zeta) -\zeta) \geq 0, && {\bf T}_r^{c+} \in \mathbb{R}_+^M .
 \end{align}
In addition both the full and reduced order activation functions satisfy the sector constraint, but the complementarity constraint is more general so the sector constraint is redundant and thus not included. Finally, the full and reduced order activation functions satisfy the cross constraints:
\begin{align}
 2 (\phi(\xi)-\xi)^T {\bf T}^{\times} \phi(\zeta) \geq 0, \quad {\bf T}^{\times} \in \mathbb{R}^{N \times M} ,\\
2 (\phi(\zeta)-\zeta)^T {\bf T}_r^{\times} \phi(\xi) \geq 0, \quad {\bf T}_r^{\times} \in \mathbb{R}^{M \times N} .
\end{align}
}
These constraints can be combined as shown in equation (\ref{eq:step}) and then, using the definitions of $\xi,\zeta,\phi(\xi)$ and $\phi(\zeta)$ given earlier, can be expressed as
given in equation (\ref{eq:step2}). Note that the slope constraints of $\text{ReLU}(\cdot)$ could have also been exploited as well. 
\begin{figure*}
\begin{align} \label{eq:step}
 \begin{bmatrix}
  \xi \\ \zeta \\ \phi(\xi) \\ \phi(\zeta) \\ 1
 \end{bmatrix}^T
\begin{bmatrix}
 0_{N \times N} & 
 0_{N \times M} & 
 {\bf T}^{0} &
 -{\bf T}^{\times} & 
 -{\bf T}_{c+} \\
 \star &
 0_{M \times M} & 
 -{\bf T}_r^{\times} &
 {\bf T}^{0}_r & 
 -{\bf T}_r^{c+} \\
 \star &
 \star & 
 -2{\bf T}^{0}  &
 {\bf T}^{\times}+({\bf T}^{\times}_r)^T &
 {\bf T}^{+} + {\bf T}^{c+} \\
 \star & 
 \star & 
 \star &
 - 2 {\bf T}^{0}_r   & {\bf T}^{+}_r + {\bf T}^{c+}_r \\
 \star & \star & \star & \star & 0
\end{bmatrix}
 \begin{bmatrix}
  \xi \\ \zeta \\ \phi(\xi) \\ \phi(\zeta) \\ 1
 \end{bmatrix} \geq 0
 \end{align}
 \begin{align}\label{eq:step2}
  \mu (x)^T 
  \underbrace{\begin{bmatrix}
   0 & 
   W_0^T {\bf T}^{0} -\Psi_0' {\bf T}^{\times}_r &
   \Psi_0^T {\bf T}_r^{0} - W_0^T {\bf T}^{\times} & 
   -W_0^T {\bf T}^{c+} - \Psi_0^T {\bf T}^{c+}_r 
    \\
        \star &
    -2 {\bf T}^{0} +{\bf T}^{0} W - W^T {\bf T}^{0} &
    (I_N-W)^T {\bf T}^{\times} + ({\bf T}_r^{\times})^T (I_M-\Psi)
    & {\bf T}^{0} b + {\bf T}^{+} + (I_N-W)^T {\bf T}^{c+} -({\bf T}^{\times}_r)^T \beta \\
    \star & 
    \star & 
    -2 {\bf T}^{0}_r +{\bf T}^{0}_r \Psi + \Psi^T {\bf T}^{0}_r & 
    {\bf T}^{0}_r \beta + {\bf T}^{+}_r + (I_M-\Psi)^T {\bf T}^{c+}_r -({\bf T}^{\times})^T b \\
    \star & \star & \star &
    -({\bf T}^{c+})^T b - b^T {\bf T}^{c+} 
    -({\bf T}^{c+}_r)^T \beta - \beta^T {\bf T}^{c+}_r
  \end{bmatrix}}_{\Lambda_{\text{ReLU}}}
  \mu(x) \geq 0
 \end{align}

 \hrule

\end{figure*}
%
%
%
%
%
%

 \subsection*{Appendix 2: Proof of Theorem \ref{thm:result}}

Inequality \eqref{qi_thm} can be split into three components:
 \begin{equation}
\resizebox{0.48\textwidth}{!}{$
  \underbrace{\begin{bmatrix}x \\ 1 \end{bmatrix}^T P_{\mathcal{X}_{\infty}}\begin{bmatrix}x \\ 1 \end{bmatrix}}_{\theta_x}  + \underbrace{\mu(x)^T \Lambda \mu(x)}_{\theta_{\phi}} 
 + \underbrace{\mu(x)^T (L^T L - \Gamma) \mu(x)}_{\theta_{\|f-g\|_2^2}} \leq 0.$}
 \end{equation}
 If $x \in \mathcal{X}_\infty$, then $\theta_x \geq 0$ from Lemma \ref{def:x_gen}. Also, if the nonlinear activation functions satisfy the quadratic constraint of Lemma \ref{ass_phi}, then $\theta_{\phi} \geq 0$. Thus, inequality (\ref{qi_thm}) implies 
 \[
  \mu(x) (L^T L - \Gamma) \mu(x) \leq 0
 \]
which then implies
 the error bound $\|f(x)-g(x)\|_2^2 \leq \gamma_x \|x\|_2^2 - \gamma$ holds for all $x \in \mathcal{X}_\infty$.

 \subsection*{Appendix 3: Proof of Proposition \ref{prop:result}}
{
Theorem \ref{thm:result} requires the following matrix inequality to hold
\[
\Pi_\infty + \Lambda   -\Gamma + L^T L \leq 0.  
\]
Using the Schur complement, a sufficient condition for this to hold is
\[
 \Omega_{\rm mod} = \begin{bmatrix}
 \Pi_\infty + \Lambda  -\Gamma & L^T \\
 L & -I 
 \end{bmatrix} \leq 0.
\]} 
This matrix is not linear due to the fact that the matrix
$\Lambda ({\bf T}^i,{\bf T}_r^i,\Psi,\Psi_0,\beta)$ contains products of the constraint matrix variables ${\bf T}^i$ and the reduced-order network paraters $\Psi, \Psi_0$ and $\beta$. However, $\Lambda$ can be re-written as in equation (\ref{eq:Lambda}) where some of the matrix variables  ${\bf T}_r^i$ are written as products of other matrix variables and two constant matrices $J_1 \in \mathbb{R}^{N \times M}$ and $J_2 \in \mathbb{R}^M$ which are chosen by the user. In equation (\ref{eq:Lambda}), the $\tilde{\Lambda}_{ij}$ elements are affine functions of the ${\bf T}^i,{\bf T}_r^i$ matrix variables and $U_{k}$ are matrix variables constructed from the sum of one or more ${\bf T}_r^i$.  
Defining
\[
 {\bf F}_0 = U_1 \Psi_0, \quad
 {\bf F}_\Psi   = U_2 \Psi, \quad
 {\bf F}_\beta = U_3 \beta,
\]
and using the expression for $\Lambda$ in $\Omega_{\rm schur}$ yields the linear matrix inequality of (\ref{eq:schur}).

Once inequality (\ref{eq:schur}) is satisfied, the parameters of the reduced order network can be recovered via
\[
 \Psi_0 = U_1^{-1} {\bf F}_0, \quad
 \Psi   = U_2^{-1} {\bf F}_\Psi, \quad
  \beta = U_3^{-1} {\bf F}_\beta.
\]

\begin{figure*}
\begin{align} \label{eq:Lambda}
 \Lambda = 
 \begin{bmatrix}
  0 
  & \tilde{\Lambda}_{12} + \Psi_0^T U_1 J_1^T
  & \tilde{\Lambda}_{13} + \Psi_0^T U_1
  & \tilde{\Lambda}_{14} + \Psi_0^T U_1 J_2  \\
  \star
  & \tilde{\Lambda}_{22} 
  & \tilde{\Lambda}_{23} + J_1 U_2 \Psi 
  & \tilde{\Lambda}_{24} + J_1 U_3 \beta \\
  \star 
  & \star
  & \tilde{\Lambda}_{33} + U_2 \Psi + \Psi^T U_2 
  & \tilde{\Lambda}_{34} +  \Psi^T U_2 J_2 + U_3 \beta \\
  \star
  & \star
  & \star 
  & \tilde{\Lambda}_{44} + \beta^T U_3 J_2 + J_2^T U_3 \beta
 \end{bmatrix}
\end{align}
\begin{align} \label{eq:schur}
\Omega_{\rm schur} =
\begin{bmatrix}
 -\tau_{x_{\infty}} -\gamma_x I_{n_x} 
  & \tilde{\Lambda}_{12} + {\bf F}_0^T J_1^T
  & \tilde{\Lambda}_{13} + {\bf F}_0^T 
  & \tilde{\Lambda}_{14} + {\bf F}_0^T J_2
  + \frac{\tau_{x_\infty}}{2} (\underbar{x}+\bar{x}) 
  & 0_{n_x \times n_f}
  \\ \star
  & \tilde{\Lambda}_{22} 
  & \tilde{\Lambda}_{23} + J_1 {\bf F}_\Psi 
  & \tilde{\Lambda}_{24} + J_1 {\bf F}_\beta 
  & (W^f)^T 
  \\
  \star 
  & \star
  & \tilde{\Lambda}_{33} + {\bf F}_\Psi + {\bf F}_\Psi^T  
  & \tilde{\Lambda}_{34} +  {\bf F}_\Psi^T  J_2 + {\bf F}_\beta 
  & -(\Psi^f)^T  
  \\
  \star
  & \star
  & \star 
  & \tilde{\Lambda}_{44} + {\bf F}_\beta^T J_2 + J_2^T {\bf F}_\beta -\gamma - \underbar{x}^T \tau_{x_{\infty}} \bar{x}
  & b^l - \beta^{\lambda} \\
  \star
  & \star
  & \star 
  & \star
  & -I_{n_f}
\end{bmatrix}
\end{align}
\hrule

\end{figure*}

 \subsection*{Appendix 4: The matrix $\Lambda$ for the case $\phi(y) = $ReLU$(y)$}

When the activation functions of the neural network are $\phi(y) =$ ReLU$(y)$, then the matrix $\Lambda=\Lambda_{\text{ReLU}}$ in $\Omega_{\text{schur}}$ is given as in inequality (\ref{eq:step2}). 
$\Omega_{\text{schur}}$ becomes an LMI if $\Lambda$ can be made linear. 
$\Lambda_{\text{ReLU}}$ of \eqref{eq:step2} features the matrix variables   ${\bf T}^{0} \in \mathbb{D}^{N}$, 
${\bf T}^{0}_r \in \mathbb{D}^{M}$, 
 ${\bf T}^{+}\in \mathbb{R}^{N}_+$, 
 ${\bf T}^{+}_{r} \in \mathbb{R}^{M}_+$, 
 ${\bf T}^{c+} \in \mathbb{R}^{N}_+$,
  ${\bf T}^{c+}_{r} \in \mathbb{R}^{M}_+$, 
  ${\bf T}^{\times} \in \mathbb{R}^{N \times M}_+$,
  ${\bf T}^{\times}_{r} \in \mathbb{R}^{N \times M}_+$, as well as
  $\Psi \in \mathbb{R}^{M \times M}$, $\Psi_0 \in \mathbb{R}^{M \times n_x}$ and $\beta \in \mathbb{R}^{N}$. 
  To make $\Lambda_{\text{ReLU}}$,  specific structures for ${\bf T}^{\times}_r$ and ${\bf T}_r^{c+}$ must be chosen, viz,
  \begin{align} \label{eq:T}
   {\bf T}^{\times}_r  =  {\bf T}_r^{0} J_1^T,  \quad
   {\bf T}_r^{c+}      =  {\bf T}_r^{0}  J_2 ,
  \end{align}
  where $J_1 \in \mathbb{R}^{N \times M}$ and $J_2 \in \mathbb{R}^M$, which makes the substitutions
\begin{align} \label{eq:F}
{\bf F}_\Psi = {\Psi}^T {\bf T}_r^{0}, \quad  {\bf F}_0 = \Psi_0 {\bf T}_r^0, \quad {\bf F}_\beta = {\beta}^T {\bf T}_r^{0}.
\end{align}
The arising expression for $\Lambda_{\text{ReLU}}$ is shown in equation (\ref{eq:relu}) and is clearly linear in the matrix variables $T^0,{\bf T}_r^{0},T^+,T_r^+,{\bf T}^{c+},{\bf T}^{\times},{\bf F}_\Psi,{\bf F}_0$ and ${\bf F}_\beta$. As in the general case, the reduced order parameters can be determined via
\[
 \Psi_0 =({\bf T}_r^{0})^{-1} {\bf F}_0, \quad
 \Psi   = ({\bf T}_r^{0})^{-1} {\bf F}_\Psi, \quad
  \beta = ({\bf T}_r^{0})^{-1} {\bf F}_\beta.
\]

The scaling matrices $J_p$, $p\in \{1,\,2\}$ are constant matrices that can be picked by the user, under the stipulation that they preserve the properties of the multipliers of Lemma \ref{ass_phi}. In this work, the choice was
\begin{align}\label{J}
J_1 = \begin{bmatrix} I_{M} \\ \bm{0}_{N-M \times M} \end{bmatrix}, ~ J_2 = \bm{1}_M,
\end{align} 
but more refined choices may also exist.

In this way, the non-convexity of the bilinear matrix inequality of the problem has been relaxed into a convex linear one. However, the substitution \eqref{eq:F} limits the space of solutions that can be searched over by the synthesis SDP, resulting in only local optima being achieved and increased conservatism in the approximation error bounds. 

\begin{figure*}
 \begin{align} \label{eq:relu}
  \Lambda_\text{ReLU} =
  \begin{bmatrix}
   0 & 
   W_0^T {\bf T}^{0} -{\bf F}_0' J_1' &
   {\bf F}_0^T - W_0^T {\bf T}^{\times} & 
   W_0^T {\bf T}^{c+} - {\bf F}_0^T J_2 
    \\
        \star &
    - 2 {\bf T}^{0} +{\bf T}^{0} W - W' ({\bf T}^{0})^T &
    (I_N-W)^T {\bf T}^{\times} + J_1 {\bf T}_r^{0} - J_1 {\bf F}_\Psi
    & {\bf T}^{0} b + {\bf T}^{+} + (I-W)^T {\bf T}^{c+} -J_1 {\bf F}_\beta \\
    \star & 
    \star & 
    - 2{\bf T}^{0}_r +{\bf F}_\Psi + {\bf F}_\Psi^T  & 
    {\bf F}_\beta + {\bf T}^{+}_r + {\bf T}_r^0 J_2 -{\bf F}_\Psi' J_2 -({\bf T}^{\times})^T b \\
    \star & \star & \star &
    -({\bf T}^{c+})^T b - b^T {\bf T}^{c+} 
    -J_2^T {\bf F}_\beta - {\bf F}_\beta^T J_2
  \end{bmatrix}
 \end{align}
\hrule
\end{figure*}

 \begin{remark}
  The use of the specific structures presented in (\ref{eq:T}) it is vital to ensure the orginal properties of the matrices are satisfed. For instance ${\bf T}^{\times}_r$ is required to have all of its elements positive (or zero). This will indeed be the case if the element of $J_1 \in \mathbb{R}_+^{N \times M}$ and if ${\bf T}_r^0 \in \mathbb{D}_{+}^{M}$ as required by Lemma \ref{lem:prop}. However, to recover the reduced order network parameters ${\bf T}_r^0$ also needs to be nonsingular, so in the arising optimisation problem it may be prudent to choose ${\bf T}_r^0$ to be strictly positive definite to guarantee this. Similar comments apply to the the vector $J_2$. 
 \end{remark}

 
 \subsection*{Appendix 5: Some remarks about Proposition \ref{prop:result}}

\paragraph{Neural network synthesis} A key feature of Proposition \ref{prop:result} is that the parameters of the reduced-order neural network are synthesized in one shot from the solution to \eqref{omega}. Directly minimising the worst-case approximation error of the reduced-order neural networks may lead to more robust and reliable out-of-sample performance. 

\paragraph{Computational cost} The main source of computational complexity in Proposition \ref{prop:result} is the growth in the number of decision variables as the problem involves matrix variables.  This limits the applicability of the proposed approach to modestly size networks. However, this issue could be reduced by imposing sparsity on the various matrix variables, such as restricting the scaling $\Lambda_p$ matrices in Lemma \ref{ass_phi} to be diagonal or sparse. Scalability issues are a common curse of methods providing robustness guarantees, like \cite{pappas}, but methods are being developed to alleviate these issues, e.g. \cite{dathathri2020enabling}.

\paragraph{Bilinearity} The BMI constraint in \eqref{eq:relu} is the source of non-convexity in the problem  \cite{bmi} which had to be relaxed. It is highly likely that, for a given full-order  neural network, there would exist better substitutions than \eqref{J}, however, \eqref{J} seemed to work quite well in the numerical example of Section \ref{sec:examples}. 

\paragraph{Robust approximation} Since the robustness analysis holds for all nonlinear activation functions satisfying the quadratic inequalities of Lemma \ref{lem:prop} and all inputs $x\in \mathcal{X}_\infty$, the performance guarantees of Proposition \ref{prop:result} may be conservative. 

\end{document}